\documentclass[10pt,twocolumn,a4paper]{ICCAS2024}
 
\usepackage{diagbox}
\usepackage{url}            
\usepackage{booktabs}       
\usepackage{amsfonts}       
\usepackage{nicefrac}       
\usepackage{microtype}      
\usepackage{color}          
\usepackage{graphicx}
\usepackage{booktabs}
\usepackage{caption}
\usepackage{float}
\usepackage{hyperref}
\usepackage[utf8]{inputenc}
\usepackage{algpseudocode}
\usepackage{algorithm}
\usepackage{multirow}

\usepackage{geometry}

\newcommand{\mytoprule}{\specialrule{0.12em}{0em}{0em}}
\newcommand{\mymidrule}{\specialrule{0.05em}{0em}{0em}}
\newcommand{\mybottomrule}{\specialrule{0.12em}{0em}{0em}}

\begin{document}

\title{Kolmogorov-Arnold Networks for Online Reinforcement Learning}

\author{Victor A. Kich${}^{1*}$, Jair A. Bottega${}^{1*}$, Raul Steinmetz${}^{2}$, Ricardo B. Grando${}^{3}$, Ayano Yorozu${}^{1}$ and Akihisa Ohya${}^{1}$}

\affils{${}^{1}$Intelligent Robot Laboratory, University of Tsukuba,\\
Tsukuba, Japan (victorkich98@gmail.com, jairaugustobottega@gmail.com) {\small${}^{*}$ Corresponding author}\\
${}^{2}$Technology Center, Federal University of Santa Maria,\\
Santa Maria, Brazil (rsteinmetz@inf.ufsm.br) \\
${}^{3}$Robotics and AI Lab, Technological University of Uruguay,\\
Rivera, Uruguay (ricardo.bedin@utec.edu.uy)}

\abstract{
Kolmogorov-Arnold Networks (KANs) have shown potential as an alternative to Multi-Layer Perceptrons (MLPs) in neural networks, providing universal function approximation with fewer parameters and reduced memory usage. In this paper, we explore the use of KANs as function approximators within the Proximal Policy Optimization (PPO) algorithm. We evaluate this approach by comparing its performance to the original MLP-based PPO using the DeepMind Control Proprio Robotics benchmark. Our results indicate that the KAN-based reinforcement learning algorithm can achieve comparable performance to its MLP-based counterpart, often with fewer parameters. These findings suggest that KANs may offer a more efficient option for reinforcement learning models. Our implementations can be found in the following link: \url{https://github.com/victorkich/Kolmogorov-PPO}.
}


\maketitle


\section{Introduction}\label{intro}

Kolmogorov-Arnold Networks (KANs) \cite{liu2024kan} are learnable approximators based on the Kolmogorov-Arnold representation theorem~\cite{kolmogoroff1957representation}, which states that any multivariate continuous function can be expressed as a finite sum of continuous univariate functions. KANs can approximate functions with fewer neurons compared to Multi-Layer Perceptrons (MLPs), thereby reducing network complexity. This simplification enables easier function decomposition, allowing for more straightforward and targeted architecture designs tailored to specific problems. Additionally, KANs utilize less memory and enhance model interpretability, making them highly appealing for various machine learning applications.

Reinforcement Learning (RL)\cite{sutton2018reinforcement} has proven efficient for continuous control tasks such as autonomous robot manipulation\cite{nguyen2019review}, navigation~\cite{zhu2021deep}, and video game playing~\cite{hafner2023mastering}, \cite{wang2023voyager}. Proximal Policy Optimization (PPO)\cite{schulman2017proximal} is a Deep RL algorithm that employs model-free online reinforcement learning to approximate an optimal policy over online training experiences. This algorithm is a cornerstone in the field, excelling in numerous benchmarks and serving as the foundation for newer models like Multi-Agent PPO~\cite{yu2022surprising}, PPO with Covariance Matrix Adaptation~\cite{hamalainen2020ppo}, PPO with Action Mask~\cite{tang2020implementing}, and PPO with Policy Feedback~\cite{gu2021proximal}.

\begin{figure}[tbp]
    \centering
    \includegraphics[width=\linewidth]{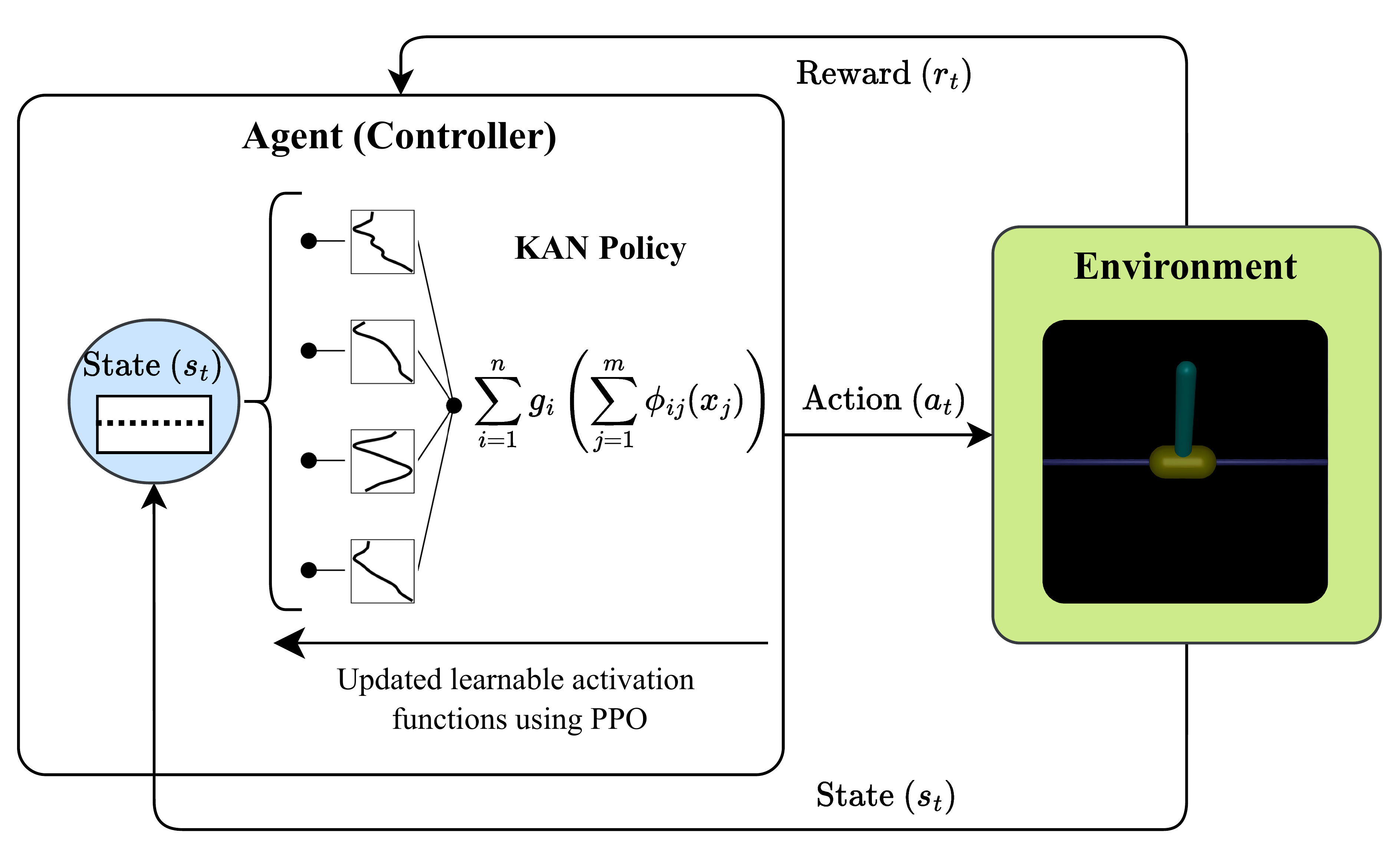}
    \caption{Overview of the proposed framework.}
    \label{fig:curled-dreamer}
    \vspace{-4mm}
\end{figure}

Networks with fewer parameters yet equal or superior approximation capabilities have the potential to significantly enhance performance in various deep reinforcement learning applications. In this work, we propose the use of KANs as policy and value approximators for PPO. The main contributions of this paper are:

\begin{itemize}
\item The first application of KANs as function approximators in a reinforcement learning algorithm.
\item A performance comparison between KAN-based and MLP-based Proximal Policy Optimization in robotics control.
\end{itemize}

The paper is organized as follows: In Section~\ref{related}, we review related works and research in the field. Section~\ref{theoretical} provides the mathematical background of the techniques employed. The proposed methodology is presented in Section~\ref{methodology}. Section~\ref{results} presents the results and evaluation analysis. Finally, we discuss our findings in Section~\ref{conclusion}.

\section{Related Works}\label{related}

KANs have recently garnered significant attention for their versatility and enhanced performance across various applications. Numerous studies have explored KANs as alternatives to traditional Multilayer Perceptrons, consistently demonstrating their superiority in specific scenarios. Concurrently, online reinforcement learning has been a cornerstone in the domain of continuous control tasks, underscoring the importance of investigating KANs' potential in this field. Evaluating whether KANs can compete with or surpass MLPs in reinforcement learning applications is a critical area of exploration. This section reviews pivotal research contributions relevant to this research.

Expanding on the KAN framework, Bozorgasl and Chen~\cite{bozorgasl2024wav} introduced Wav-KAN, a neural network architecture that integrates wavelet functions~\cite{antoniadis1997wavelets} into the KAN framework. This innovation addresses common challenges faced by traditional MLPs, such as interpretability, training speed, robustness, computational efficiency, and performance. By incorporating wavelet functions, Wav-KAN effectively captures both high-frequency and low-frequency components of the input data, enhancing overall performance.

Exploring KANs' performance on predicting time series, Rubio et al.~\cite{vaca2024kolmogorov} demonstrated the effectiveness of these networks in predicting satellite traffic, showing their superiority over traditional MLPs for time series analysis. Similarly, Genet and Hugo~\cite{genet2024tkan} introduced Temporal Kolmogorov-Arnold Networks (TKANs), combining elements of KANs and Long Short-Term Memory (LSTM) networks to improve time series forecasting.

Expanding on the exploration of KANs' capabilities, Yang, Qin, and Yu~\cite{yang2024endowing} proposed a novel framework that enhances the interpretability of neural cognitive diagnosis models (CDMs) using KANs while maintaining high accuracy. They introduced two methods: replacing traditional MLPs in existing neural CDMs with KANs and designing a new aggregation framework entirely based on KANs. Their experimental results on four real-world datasets showed that the KA2NCD framework outperforms traditional and neural CDMs in both accuracy and interpretability, demonstrating its potential in intelligent education systems.

Online Reinforcement Learning, where an agent continuously learns from ongoing interaction with the environment, instead of using memory replay, has been successfully applied to various robotic tasks. 
In the humanoid robots area, Melo et al.~\cite{melo2021learning} successfully utilized PPO to develop running skills in a humanoid robot, using the RoboCup 3D Soccer Simulation environment, where humanoid robots compete in simulated soccer matches. Expanding the use of online reinforcement learning to other areas, Proctor~\cite{proctor2021proximal} used the algorithm for localizing and searching for lost nuclear sources in a designated area, which is vital for societal safety and reducing human harm.

To the best of our knowledge, no efforts have been made to apply KANs as alternatives to MLPs in online reinforcement learning, making this study a pivotal contribution.

\section{Theoretical Background}\label{theoretical}

This section presents the theory and mathematical foundations underpinning our approach.

\subsection{Kolmogorov-Arnold Networks}

Kolmogorov-Arnold Networks leverage the Kolmogorov-Arnold representation theorem, which asserts that any multivariate continuous function can be decomposed into a finite composition of univariate functions and addition operations. Formally, for a smooth function \( f: [0,1]^n \to \mathbb{R} \), this can be expressed as:

\begin{equation}
f(x) = \sum_{q=1}^{2n+1} \Phi_q \left( \sum_{p=1}^{n} \varphi_{q,p}(x_p) \right),
\end{equation}

\noindent where \( \varphi_{q,p}: [0,1] \to \mathbb{R} \) and \( \Phi_q: \mathbb{R} \to \mathbb{R} \) are continuous functions.

In KANs, weight parameters are replaced by learnable 1D functions \( \varphi \), parametrized as B-splines. The computation in a KAN layer with \( n_{\text{in}} \) inputs and \( n_{\text{out}} \) outputs is:

\begin{equation}
x_{l+1,j} = \sum_{i=1}^{n_l} \varphi_{l,j,i}(x_{l,i}),
\end{equation}

\noindent where \( \varphi_{l,j,i} \) is a spline function connecting the \( i \)-th neuron in layer \( l \) to the \( j \)-th neuron in layer \( l+1 \).

The backpropagation process in KANs involves calculating gradients of the spline functions. The loss \( \mathcal{L} \) is minimized using gradient descent, with the gradient of the loss with respect to the spline parameters \( c_i \) computed as:

\begin{equation}
    \frac{\partial \mathcal{L}}{\partial c_i} = \sum_{j=1}^{n_{\text{out}}} \frac{\partial \mathcal{L}}{\partial x_{l+1,j}} \frac{\partial x_{l+1,j}}{\partial c_i},
\end{equation}

\noindent where \( \frac{\partial x_{l+1,j}}{\partial c_i} \) involves the derivative of the spline function with respect to its coefficients.

\subsubsection{Grid and Order in KANs}

In KANs, the spline functions \( \varphi \) are defined over a discretized domain, commonly referred to as a \textit{grid}. The \textbf{grid} determines the points at which the spline functions are evaluated, essentially specifying the resolution of the function approximation. The \textbf{order} parameter controls the degree of the B-splines used to represent the functions. For example, an order of 1 corresponds to linear splines, while higher orders correspond to splines with more complex shapes, allowing for more flexible function approximations.

The combination of the \textit{grid} and \textit{order} parameters defines the complexity and capacity of the KANs to model intricate functional relationships. A finer grid with higher-order splines enables more accurate approximations of complex functions, at the cost of increased computational complexity.

\subsubsection{Comparison of KANs and MLPs}

\begin{itemize}
    \item \textbf{Learnable Activation Functions:} KANs feature learnable activation functions on edges, while MLPs have fixed activation functions on nodes.
    \item \textbf{Spline-Based Weights:} KANs use splines to represent weights, enhancing their capacity to approximate complex functions with fewer parameters.
    \item \textbf{Layer Extension:} Both KANs and MLPs can be extended to multiple layers.
\end{itemize}

\subsection{Proximal Policy Optimization}

Proximal Policy Optimization is a policy gradient method that computes an estimator of the policy gradient for use in a stochastic gradient ascent algorithm. The policy gradient estimator is:
\begin{equation}
\hat{g} = \hat{\mathbb{E}}_t \left[ \nabla_\theta \log \pi_\theta(a_t | s_t) \hat{A}_t \right],
\end{equation}
\noindent where \(\pi_\theta\) is the stochastic policy, and \(\hat{A}_t\) is an estimator of the advantage function at timestep \( t \). The expectation \(\hat{\mathbb{E}}_t[ \cdots ]\) denotes the empirical average over a finite batch of samples.

This estimator is derived from the objective function:
\begin{equation}
L_{PG}(\theta) = \hat{\mathbb{E}}_t \left[ \log \pi_\theta(a_t | s_t) \hat{A}_t \right].
\end{equation}

To prevent excessively large policy updates, PPO introduces a clipped surrogate objective. Let \( r_t(\theta) \) denote the probability ratio:
\begin{equation}
r_t(\theta) = \frac{\pi_\theta(a_t | s_t)}{\pi_{\theta_{\text{old}}}(a_t | s_t)}.
\end{equation}

The clipped surrogate objective is defined as:
\begin{equation}
\footnotesize
L_{CLIP}(\theta) = \hat{\mathbb{E}}_t \left[ \min \left( r_t(\theta) \hat{A}_t, \text{clip}(r_t(\theta), 1 - \epsilon, 1 + \epsilon) \hat{A}_t \right) \right],
\end{equation}
\noindent where \(\epsilon\) is a hyperparameter, typically set to 0.2. This objective encourages the new policy to stay within a small region around the old policy by clipping the probability ratio \( r_t(\theta) \) when it exceeds the interval \([1 - \epsilon, 1 + \epsilon]\).

The advantage function \(\hat{A}_t\) can be estimated using Generalized Advantage Estimation (GAE), which reduces variance while introducing a bias controlled by parameter \(\lambda\):
\begin{equation}
\hat{A}_t = \sum_{l=0}^{\infty} (\gamma \lambda)^l \delta_{t+l},
\end{equation}
\noindent where \(\delta_t\) is the temporal difference (TD) error:
\begin{equation}
\delta_t = r_t + \gamma V(s_{t+1}) - V(s_t).
\end{equation}

\begin{algorithm}[bp]
\caption{Proximal Policy Optimization}
\label{algorithm}
\begin{algorithmic}[1]
\State Initialize policy parameters \(\theta\)
\For{each iteration}
    \For{actor = 1, 2, ..., N}
        \State Run policy \(\pi_\theta\) in environment for \(T\) timesteps
        \State Compute advantage estimates \(\hat{A}_t\)
    \EndFor
    \State Optimize surrogate \(L_{\text{CLIP}}(\theta)\) with \(K\) epochs and minibatch size \(M \leq NT\)
    \State Update policy parameters \(\theta\)
\EndFor
\end{algorithmic}
\end{algorithm}

In practice, PPO uses a combination of the clipped surrogate objective, a value function loss, and an entropy bonus to ensure sufficient exploration. The combined objective function is:
\begin{equation}
\footnotesize
L_{CLIP+VF+S}(\theta) = \hat{\mathbb{E}}_t \left[ L_{CLIP}(\theta) - c_1 L_{VF}(\theta) + c_2 S[\pi_\theta](s_t) \right],
\end{equation}
\noindent where \( L_{VF}(\theta) = (V_\theta(s_t) - V_t^\text{target})^2 \) is the squared-error loss for the value function, \( S[\pi_\theta](s_t) \) is the entropy bonus, and \( c_1, c_2 \) are coefficients.

The PPO technique (Algorithm~\ref{algorithm}) alternates between collecting data through interaction with the environment and optimizing the policy using the clipped surrogate objective.

\section{Methodology}\label{methodology}

We conducted our experiments using six environments from the Gymnasium Mujoco suite, specifically adapted from the DeepMind Control Proprio benchmarks \cite{tassa2018deepmind}. These environments are well-suited for evaluating continuous control tasks and include: HalfCheetah-v4, Hopper-v4, InvertedPendulum-v4, Pusher-v4, Swimmer-v4, and Walker2d-v4. These environments provide a diverse set of challenges, testing the ability of the agent to learn different types of motion and control strategies.

Our implementation utilizes the PPO algorithm as the core reinforcement learning method. PPO is a state-of-the-art algorithm known for its stability and efficiency in policy gradient methods \cite{schulman2017proximal}. We incorporated KANs as the function approximators for both the policy (actor) and value (critic) functions, replacing the traditional MLP used in standard PPO implementations.

\subsection{Network Architectures}

We designed and compared several network architectures for the policy (actor) and value (critic) functions to evaluate the effectiveness of KANs versus MLPs. The configurations included:
\begin{itemize}
\item \textbf{MLP (a=2, c=2)}: Two hidden layers for both actor and critic with 64 units each.
\item \textbf{MLP (a=1, c=2)}: One hidden layer for the actor and two hidden layers for the critic, with 64 units each.
\item \textbf{KAN (k=2, g=3)}: KAN for the actor with no hidden layers, order $k=2$ and grid $g=3$, and a critic implemented as an MLP with 2 hidden layers of 64 units each.
\item \textbf{Full KAN (k=2, g=3)}: Both the actor and critic networks implemented using KANs with no hidden layers, order $k=2$ and grid $g=3$.
\end{itemize}

The parameter settings for KANs were chosen based on preliminary tests where we evaluated various configurations to identify the most effective setup. Our tests demonstrated that a KAN with no hidden layers, order $k=2$, and grid $g=3$ provided the best performance in terms of computational efficiency and approximation capability. Consequently, we adopted this architecture for intensive testing.

\begin{table}[bp]
  \vspace{-5mm}
  \caption{Number of parameters (actor) for different DMC Proprio environments.}
  \label{params-table}
  \centering
  \setlength{\tabcolsep}{4.5pt}
  \begin{tabular}{lrrr}
    \mytoprule
    \textbf{Task} & \textbf{MLP} & \textbf{MLP} & \textbf{KAN} \\
    \textit{Settings} & {\small(a=2, c=2)} & {\small(a=1, c=2)} & {\small(k=2, g=3)} \\
    \mymidrule
    HalfCheetah-v4       & 5702   & 1542    & 510 \\
    Walker2d-v4          & 5702   & 1542    & 510 \\
    Hopper-v4            & 5123   & 963     & 165 \\
    InvertedPendulum-v4  & 4545   & 385     & 20  \\
    Swimmer-v4           & 4866   & 706     & 80  \\
    Pusher-v4            & 6151   & 1991    & 805 \\
    \mymidrule
    \textit{Average of Parameters} & 5348 & 1188 & 348 \\
    \mybottomrule
  \end{tabular}
\end{table}

\begin{figure*}[tbp]
     \centering
     \includegraphics[width=\linewidth]{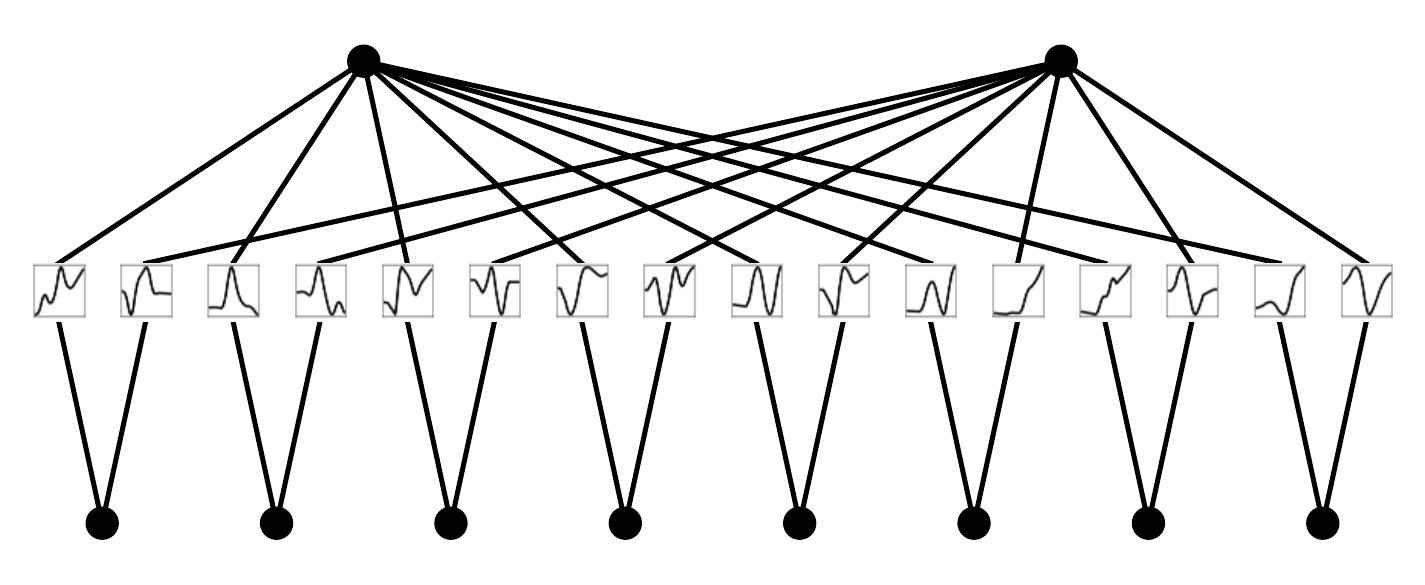}
     \captionsetup{skip=1pt}
     \caption{KAN architecture of the environment DMC Proprio Swimmer-v4.}
     \label{fig:kan_swimmer}
     \vspace{-2mm}
\end{figure*}

For the Kolmogorov-Arnold Network, we set the parameters $k=2$ and $g=3$ based on these preliminary findings, ensuring the network's ability to approximate the continuous functions involved in control tasks while maintaining low computational complexity. The number of parameters in the KAN configurations is significantly lower compared to the MLP counterparts, highlighting the efficiency of KANs. The average number of parameters for each configuration across the environments is shown in Table~\ref{params-table}. An  Kolmogorov-Arnold Network example is depicted in Fig.\ref{fig:kan_swimmer}.

\begin{figure*}[tbp]
    \centering
    \includegraphics[width=\linewidth]{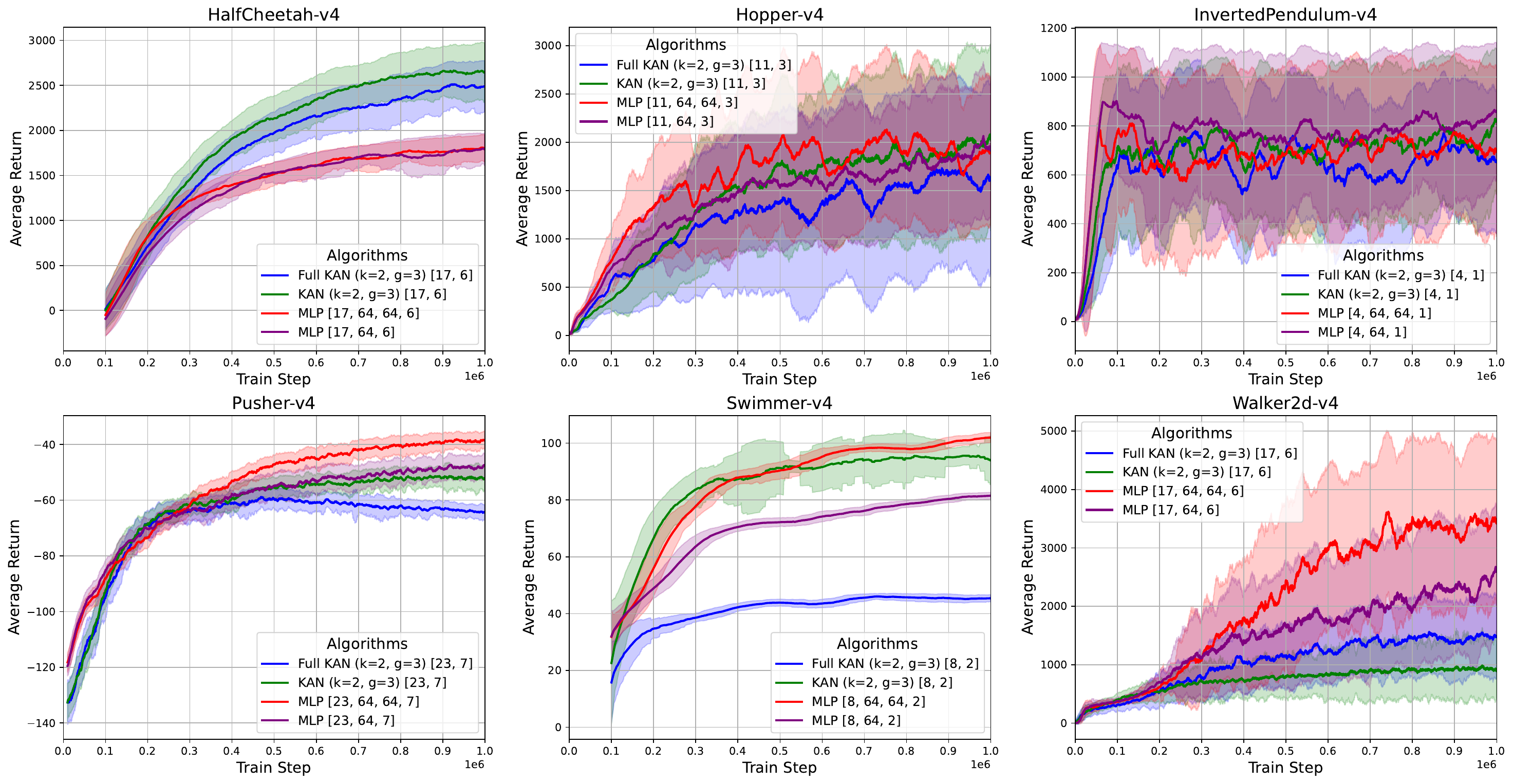}
    \caption{Reward average comparison for all the trained environments using the proposed architectures (using 5 seeds).}
    \label{fig:rewards}
\end{figure*}

\begin{table*}[tbp]
  \caption{DMC Proprio scores for control inputs at 1M train steps. Scores between the top 5\% are shown in bold format.}
  \label{dmc-scores-table}
  \centering
  \setlength{\tabcolsep}{10pt}
  \begin{tabular}{lrrrrr}
    \mytoprule
    \textbf{Task} & {\small\textbf{Full KAN (k=2, g=3)}} & {\small\textbf{KAN (k=2, g=3)}} & {\small\textbf{MLP (a=2, c=2)}} & {\small\textbf{MLP (a=1, c=2)}} \\
    \textit{Train Steps} & 1M & 1M & 1M & 1M \\
    \mymidrule
    HalfCheetah-v4      & 2490   & \textbf{2644}    & 1803   & 1791 \\
    Hopper-v4           & 1633    & \textbf{2031}   & 1897   & 1918 \\
    InvertedPendulum-v4 & 653   & \textbf{818}   & 694   & \textbf{861} \\
    Pusher-v4           & -64  & -52  & \textbf{-39}  & -47 \\
    Swimmer-v4          & 45   & 94   & \textbf{102}   & 82 \\
    Walker2d-v4         & 1487   & 889   & \textbf{3439}   & 2593 \\
    \mymidrule
    \textit{Reward Mean}       & 1358   & \textbf{1540}   & 1482   & 1399 \\
    \textit{Average of Parameters (actor+critic)} &  415  &  5490  &  10490  &  6330 \\
    \mybottomrule
  \end{tabular}
\end{table*}

\subsection{Training Procedure}

Each agent was trained for 1 million steps in each environment. We used the following hyperparameters for training:
\begin{itemize}
    \item \textbf{Number of Environments}: $1$
    \item \textbf{Learning Rate}: $3 \times 10^{-4}$
    \item \textbf{Clip Parameter $\epsilon$}: $0.2$
    \item \textbf{Number of Epochs}: 10
    \item \textbf{Batch Size}: 64
    \item \textbf{Discount Factor $\gamma$}: 0.99
    \item \textbf{GAE Parameter $\lambda$}: 0.95
    \item \textbf{Seeds}: [0, 1, 2, 3, 4]
    \item \textbf{Device}: CPU
\end{itemize}

The agents' performance was evaluated based on the average cumulative reward obtained over 100 evaluation episodes after the training period. The evaluation episodes were conducted without exploration noise to accurately assess the learned policies' effectiveness.

\section{Results}\label{results}


The performance of the KAN-based PPO was evaluated against the MLP-based PPO using metrics such as average return and the number of parameters. The average return, defined as the mean cumulative reward over 100 evaluation episodes, provided a robust measure of the policy's effectiveness. The number of parameters, representing the total count of trainable elements in the network, offered insights into the efficiency and potential computational savings achieved by using KANs.

\subsection{Performance Metrics}

Figure~\ref{fig:rewards} illustrates the average rewards obtained across the training steps for each environment. In general, the KAN-based PPO matched or exceeded the performance of the MLP-based PPO in most environments. This demonstrates the potential of KANs to achieve robust performance with fewer parameters, particularly in the task HalfCheetah-v4, where the KAN-based models outperformed their MLP counterparts.

Table~\ref{dmc-scores-table} shows the DMC Proprio scores for the control inputs at 1 million training steps. The KAN-based models performed competitively, sometimes surpassing the MLP configurations in terms of reward mean while maintaining a lower parameter count. This highlights the potential of KANs to provide efficient and effective learning with potentially reduced computational requirements.

\subsection{Discussion}

The results from our experiments indicate that KANs, when integrated into the PPO framework, can achieve performance comparable to traditional MLPs with significantly fewer parameters. This demonstrates the potential efficiency of KANs in reducing the computational complexity of reinforcement learning models without compromising performance.

While KAN-based models generally matched or exceeded the performance of MLP-based models, these improvements were context-dependent. For example, in tasks such as HalfCheetah-v4 and Hopper-v4, the KAN-based PPO not only achieved higher rewards but also maintained a lower parameter count, highlighting its efficiency. However, in environments like Pusher-v4 and Walker2d-v4, the performance gains were less pronounced or even inferior, suggesting that the benefits of KANs can vary depending on task complexity. Adjustments in parameters, grid searches, or incorporating additional neurons could potentially enhance performance in these cases.

The reduced parameter count in KANs translates to lower memory usage and potentially faster training times, making them suitable for deployment in resource-constrained environments. For instance, the KAN (k=2, g=3) architecture required significantly fewer parameters compared to the MLP (a=2, c=2) configuration, averaging 348 versus 5348 parameters, respectively. This substantial reduction underscores the efficiency of KANs in achieving comparable performance with fewer resources.

Additionally, pruning operations—where less important parameters are eliminated—can further streamline KAN models, enhancing efficiency without compromising performance. This pruning process is particularly beneficial for KANs due to their already reduced parameter count, allowing the model to focus on the most critical parameters. The inherent interpretability of KANs, based on the Kolmogorov-Arnold representation theorem, also provides advantages for understanding and debugging learned policies, which is crucial in applications where model transparency and explainability are important. However, continued exploration and optimization of KAN architectures are necessary to fully leverage their potential across various reinforcement learning applications.

KANs, while providing significant benefits in terms of parameter count and interpretability, currently face challenges in computational speed. In our evaluation of 1000 steps on the HalfCheetah environment, the MLP with 2 hidden layers of 64 units completed the task significantly faster, averaging 0.37 seconds compared to the KAN (k=2, g=3), which took 3.39 seconds. This slower performance stems from the complex calculations required by the B-splines used as activation functions, which, despite reducing the number of parameters, introduce a trade-off between parameter efficiency and execution speed—especially critical in real-time applications. MLPs, though more parameter-heavy, benefit from highly optimized implementations in libraries like Torch, resulting in faster execution times. The novelty of KANs means there is a lack of specialized optimization techniques, presenting a clear need for future research to enhance their computational efficiency.

Addressing these computational bottlenecks will be essential to fully leverage the potential of KANs in reinforcement learning tasks. Optimizing spline calculations and developing KAN-specific computational strategies could improve their competitiveness in terms of execution speed, making them more viable for applications that demand real-time decision-making while maintaining their advantages in memory usage and training efficiency. Future work should focus on overcoming these challenges to realize the full capabilities of KANs in practical settings.

\section{Conclusion}\label{conclusion}

Our study demonstrated that Kolmogorov-Arnold Networks can achieve performance comparable to traditional Multi-Layer Perceptrons with significantly fewer parameters in reinforcement learning tasks. This makes KANs a promising alternative, especially in resource-constrained environments where memory and computational efficiency are critical. However, the current limitations in computational speed due to the complexity of B-spline activation functions suggest the need for further research and optimization. Future work should focus on developing specialized optimization techniques and frameworks for KANs to improve their computational efficiency and make them more suitable for real-time applications.
\section*{ACKNOWLEDGEMENT}

This work was supported by the Ministry of Education, Culture, Sports, Science and Technology (MEXT) scholarship. We are also grateful for the support provided by the Human-Centered AI program at the University of Tsukuba.

%
\bibliographystyle{IEEEtran}
\bibliography{ref}
%

\end{document}